\pdfoutput=1

\documentclass[11pt]{article}

\usepackage[preprint]{acl}

\usepackage{times}
\usepackage{latexsym}
\usepackage{arydshln}
\usepackage[T1]{fontenc}

\usepackage{amsmath}
\usepackage{cleveref}
\usepackage{listings}
\usepackage{titlesec}
\usepackage{multirow}
\usepackage{makecell}
\usepackage{microtype}
\usepackage{hyperref}
\usepackage{graphicx} 
\usepackage{url}
\usepackage{booktabs}

\usepackage[most]{tcolorbox}

\usepackage[utf8]{inputenc}

\usepackage{microtype}

\usepackage{inconsolata}

\usepackage{graphicx}
\usepackage{subfigure}
\usepackage{enumitem}

\definecolor{demphcolor}{gray}{.45}
\newcommand{\demph}[1]{\textcolor{demphcolor}{#1}}

\NewDocumentCommand{\name}{ mO{} }{\textsc{WiNELL}}

\NewDocumentCommand{\heng}
{ mO{} }{\textcolor{red}{\textsuperscript{\textit{Heng}}\textsf{\textbf{\small[#1]}}}}
\NewDocumentCommand{\revanth}
{ mO{} }{\textcolor{brown}{\textsuperscript{\textit{Revanth}}\textsf{\textbf{\small[#1]}}}}
\NewDocumentCommand{\cheng}
{ mO{} }{\textcolor{orange}{\textsuperscript{\textit{Cheng}}\textsf{\textbf{\small[#1]}}}}
\NewDocumentCommand{\ruhi}
{ mO{} }{\textcolor{blue}{\textsuperscript{\textit{Ruhi}}\textsf{\textbf{\small[#1]}}}}

\NewDocumentCommand{\jiaxin}
{ mO{} }{\textcolor{pink}{\textsuperscript{\textit{Jiaxin}}\textsf{\textbf{\small[#1]}}}}

\title{\textsc{WiNELL}: Wikipedia Never-Ending Updating with LLM Agents}

\author{Revanth Gangi Reddy\thanks{Equal Contribution.}$^1$\hspace{0.3em}Tanay Dixit\footnotemark[1]$^1$\hspace{0.3em}Jiaxin Qin$^1$\hspace{0.3em}Cheng Qian$^1$\hspace{0.3em}Daniel Lee$^1$\\\textbf{Jiawei Han}$^1$\hspace{0.4em}\textbf{Kevin Small}$^2$\hspace{0.4em}\textbf{Xing Fan}$^2$\hspace{0.4em}\textbf{Ruhi Sarikaya}$^2$
\hspace{0.4em}\textbf{Heng Ji}$^2$\\
$^1$ University of Illinois Urbana-Champaign\hspace{1em}$^2$ Amazon \\
  \texttt{revanth3@illinois.edu; jihj@amazon.com}}

\begin{document}
\maketitle
\begin{abstract}
Wikipedia, a vast and continuously consulted knowledge base, faces significant challenges in maintaining up-to-date content due to its reliance on manual human editors. Inspired by the vision of continuous knowledge acquisition in NELL~\cite{nell2010} and fueled by advances in LLM-based agents, this paper introduces 
\name{}\footnote{Code, data and models will be available here: \url{https://github.com/gangiswag/AutoWikiUpdate}}
, an agentic framework for continuously updating Wikipedia articles. Our approach employs a multi-agent framework to aggregate online information, select new and important knowledge for a target entity in Wikipedia, and then generate precise edit suggestions for human review. Our fine-grained editing models, trained on Wikipedia's extensive history of human edits, enable incorporating updates in a manner consistent with human editing behavior. Our editor models outperform both open-source instruction-following baselines and closed-source LLMs (e.g., GPT-4o) in key-information coverage and editing efficiency. 
End-to-end evaluation on high-activity Wikipedia pages demonstrates \name{}'s ability to identify and suggest timely factual updates. This opens up a promising research direction in LLM agents for automatically updating knowledge bases in a never-ending fashion.

\end{abstract}

\section{Introduction}
\label{sec:introduction}

The visionary Never-Ending Language Learning (NELL) framework~\cite {nell2010} pioneered autonomous, continuous knowledge extraction and self-correction from web data. Though constrained by the open-domain Information Extraction capabilities of its time, NELL provided a conceptual blueprint for dynamic knowledge acquisition in intelligent systems.  Today, fueled by the rapid advancements in large language model (LLM)-based agents for information aggregation~\cite{openai2025deep, InfoAgent2025}, we are inspired to revisit and reimagine NELL's foundational ideas. 


\begin{figure}[t]
    \centering
    \includegraphics[width=0.98\linewidth]{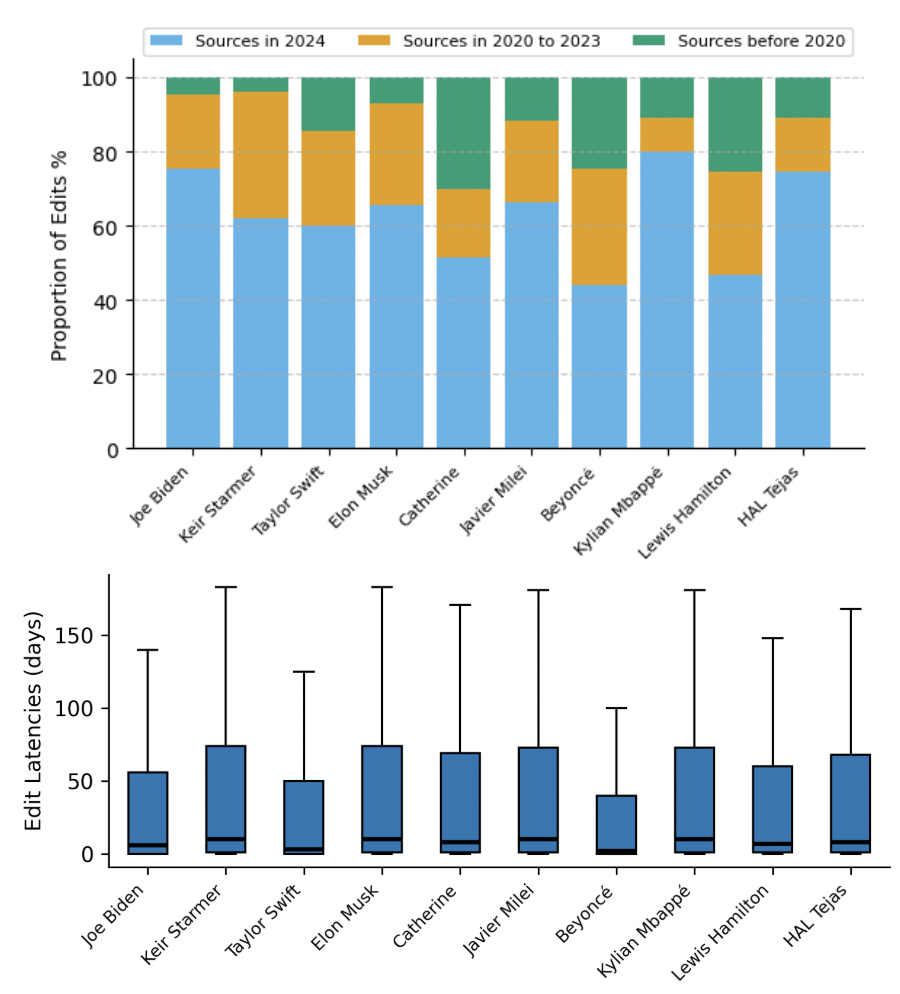}
    \vspace{-0.5em}
    \caption{Analysis of Wikipedia edits in 2024 for selected public figures. (Top) Proportion of factual updates, i.e. having citations, with cited sources from the same year.
    (Bottom) Latency distribution (days) between source publication and the subsequent Wikipedia edit, illustrating typical human update delays.}
    \label{fig:edit_latency}
    \vspace{-1em}
\end{figure}





\begin{figure*}[t]
    \centering
    \includegraphics[width=\linewidth]{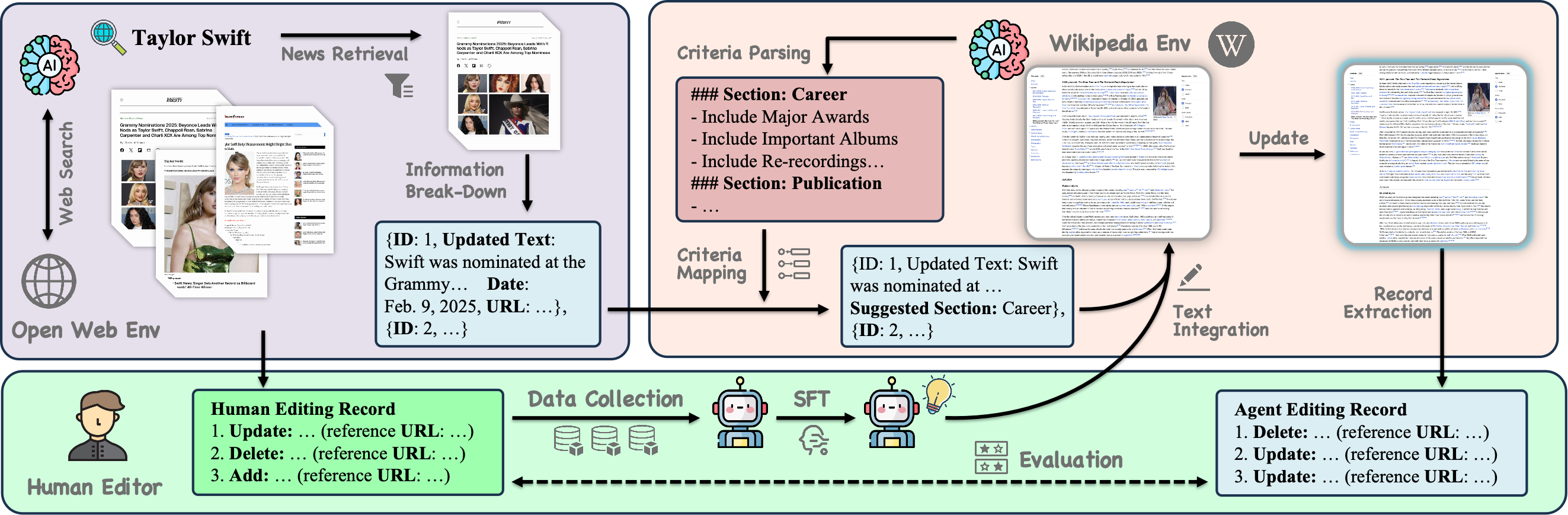}
    \vspace{-1.7em}
    \caption{Overview of the multi-stage process in \name{} for automatically updating Wikipedia articles. It first analyzes the article's structure to define section-specific content criteria, then iteratively searches the web, identifies potential updates, and aggregates relevant, non-redundant facts using an agentic framework. Finally, the editor integrates these updates into the appropriate sections, mimicking human editing patterns learnt from historical data.}
    \label{fig:method}
\end{figure*}

In this context, we present the first case study on automatic updating of Wikipedia, one of the most comprehensive and widely consulted knowledge repositories. Wikipedia faces significant challenges in maintaining up-to-date content due to its predominantly manual update process. This reliance on volunteer editors-who have collectively made over a billion edits on English Wikipedia\footnote{\url{https://en.wikipedia.org/wiki/Wikipedia:Time_Between_Edits}}-often results in substantial latency in incorporating new information. Analysis of recent edits for public figures (Figure~\ref{fig:edit_latency}) reveals considerable delays between source publication and Wikipedia updates, and many edits use sources published months or years prior. Further, less-trafficked articles frequently lag behind for extended periods~\citep{schmidt2023diachronic}.

While prior approaches have attempted to tackle the problem of automatically updating Wikipedia, they have primarily focused on infoboxes~\citep{ji2010an,ji2011an,ji2011knowledge,tompkins2012sawus, tran2013automatic, barth2023detecting,surdeanu2014overview} or assume access to the relevant facts that need to be incorporated into the update~\citep{shah2020automatic}.
Building on recent advances in agentic capabilities of large language models~\citep{liuagentbench, openai2025deep,EscapeBench2025,Mobile-Agent-E2025}, we introduce \name{}, an agentic approach to Wikipedia updating.
Given a specific Wikipedia article, \name{} continuously monitors online sources for recent facts, identifies relevant updates for the article under consideration, and automatically generates well-formed edit suggestions--complete with citations to sources.

Our approach incorporates a multi-agent framework for online information aggregation~\cite{InfoAgent2025} that iteratively searches for relevant updates for the given article and consolidates the aggregated information to ensure \name{}'s edits are precise and non-redundant. Moreover, we leverage Wikipedia's rich history of human edits to train a custom fine-grained editing model. The objective is to enable the model to integrate the identified updates into Wikipedia in a manner consistent with human editing behavior--preserving key factual information, ignoring trivial details, and maintaining objectivity. 
Figure \ref{fig:method} provides an overview of our proposed approach. 
\name{} can minimize editor workload by having humans simply review and approve suggested edits. Its human-in-the-loop design ensures quality while automating update identification and subsequent article editing. Compared to manual updates, \name{} (1) continuously ingests new information to shrink publication-to-Wikipedia edit lag, (2) can offer balanced coverage by surfacing updates across popular and overlooked topics, and (3) frees editors from time-consuming monitoring tasks so they can focus on verification and quality control.


To evaluate \name{} at scale, manual verification of generated edits is infeasible. We therefore design an automatic evaluation setup that assesses our framework's ability to cover factual updates made by human editors within a defined historical period. Specifically, we task  \name{} with suggesting edits to a Wikipedia article version at time T, using only sources published between T and T+$\Delta t$. We then compare these suggestions against factual human edits made during the same 
$\Delta t$. This involves measuring the extent to which \name{}'s edits entail the atomic facts within human edits, thereby quantifying coverage. 

In summary, our main contributions are:
\begin{itemize}[topsep=3.5pt, partopsep=5pt, leftmargin=12pt]
    \item Introduction of \name{}, an agentic framework designed to autonomously update Wikipedia articles based on online information aggregation.
    \vspace{-1.8em}
    \item Creation of a fine-grained editing model finetuned on historical human Wikipedia edits, enabling performance better than closed-source models and zero-shot variants. 
   \vspace{-0.6em}
    \item Design of an automatic evaluation setup that uses historical human edits as a benchmark to measure the performance of \name{}, quantifying the coverage of factual updates made by humans within the agent's suggestions. 
\end{itemize}

\section{Related Work}



\subsection{AI-Assisted Wikipedia Editing}

While Wikipedia has long utilized rule-based bots for narrow automated editing tasks such as data updates and vandalism reversion \citep{steiner2014bots}, recent advancements have seen the development of AI-driven systems by researchers and Wikimedia teams to aid in content maintenance, including suggesting relevant content \citep{fetahu2015automated}, detecting inconsistencies \citep{hsu2021wikicontradiction}, and recommending citations \citep{fetahu2016finding, redi2019citation, petroni2023improving}.  Previous research has also explored the generation of entire Wikipedia articles from scratch, employing methods like structure-aware template induction from existing articles and web-based content retrieval \citep{sauper-barzilay-2009-automatically}, or synthesizing topic outlines and leveraging multi-perspective question asking \citep{shao2024assisting}. Furthermore, the NIST TAC Knowledge Base Population track~\cite{ji2010an,ji2011an,ji2011knowledge,surdeanu2014overview,ji2014overview,ji2015overview,ji2017overview,Ji2019,Ji2020overview} has dedicated extensive effort to automatically populating knowledge bases, such as Wikipedia infoboxes, through entity extraction and linking. In contrast, \name{} differs fundamentally from these approaches by concentrating on \textit{updating} existing articles rather than generating new ones from scratch or tackling infoboxes, and, for the first time, adopts modern agentic LLM techniques for Wikipedia knowledge updating.


\subsection{Online Information Seeking}
Agentic approaches leveraging LLMs are increasingly employed for online information seeking through automated, iterative search processes.  Early examples, such as WebGPT~\citep{nakano2021webgpt}, involved fine-tuning models to navigate web browsers and answer open-ended questions, while prompting strategies like ReAct~\citep{yaoreact} enabled LLMs to interleave reasoning with actions such as API calls. More recent advancements feature multi-agent systems~\citep{guo2024large, tran2025multi} where AI agents with specialized roles--such as (\textit{Navigator}, \textit{Extractor}, \textit{Aggregator})~\citep{InfoAgent2025} or (\textit{Planner}, \textit{Searcher})~\citep{hu2024level, chen2024mindsearch}--collaborate to achieve complex question-answering goals. While \name{} utilizes techniques from online information seeking, its objective diverges from typical agentic question answering (QA)~\cite{krishna2024fact, wei2025browsecomp}, which aims to provide concise answers to specific user queries by retrieving and synthesizing web information. Instead, \name{} focuses on knowledge base maintenance for a given Wikipedia article, necessitating continuous and broad monitoring for any new, relevant factual developments pertinent to that article rather than addressing singular questions.


\section{\textsc{WiNELL} Methodology}
\label{sec:method}

Identifying timely, accurate, and context-appropriate facts in order to update a given Wikipedia article demands more than a one‐shot extractor or a static pipeline. Web sources evolve constantly, and relevant updates can be buried under noisy or redundant reports. An agentic aggregation process--one that reasons about where to look, how to interpret evidence, and how to refine the search strategy--is therefore essential to ensure identified updates are precise, non-redundant and have high coverage. This mirrors human editors' iterative fact‐finding: noticing gaps, testing alternative keywords or sources, and homing in on the most relevant reports~\cite{marchionini1995information, marchionini2006exploratory, thomas2024information}. 

Concretely, \name{} tackles the complex task of automatically updating a Wikipedia article as follows: A) Capturing what sections are present in the article and what kind of content is present within these sections to construct a set of informativeness criteria on the fly  (\S{\ref{sec:wiki_criteria}}), B) Iteratively searching the web to identify updates for the article under consideration (\S{\ref{sec:update_aggregation}}), C) Fine-grained article editing to incorporate these updates into the specific section that they are most relevant to (\S{\ref{sec:editing}}).


\subsection{Section Criteria Induction}
\label{sec:wiki_criteria}

Updating a Wikipedia article in a structured, coherent way hinges on understanding what belongs in each section. Different sections carry different categories of important information--e.g., `Early Life' captures biographical background, while `Professional Career' documents key milestones--so any new fact must satisfy the expectations of its target section. Hence, \name{} leverages the article's own structure--its nested hierarchy of section headings and associated content--to induce section-wise customized criteria. Specifically, we pass the entire Wikipedia article (with the section headings marked) as input to an LLM and prompt it to output a set of content inclusion criteria that specify the types of facts or updates that are important for each section in the article. The resulting criteria serve as a precise policy for the subsequent \textit{Agentic Update Aggregation} (in \S{\ref{sec:update_aggregation}}), guiding where a newly discovered fact belongs, ensuring that \name{}'s suggestions conform to the article's existing organization.

\subsection{Agentic Update Aggregation}
\label{sec:update_aggregation}
\begin{figure}
    \centering
    \includegraphics[width=\linewidth]{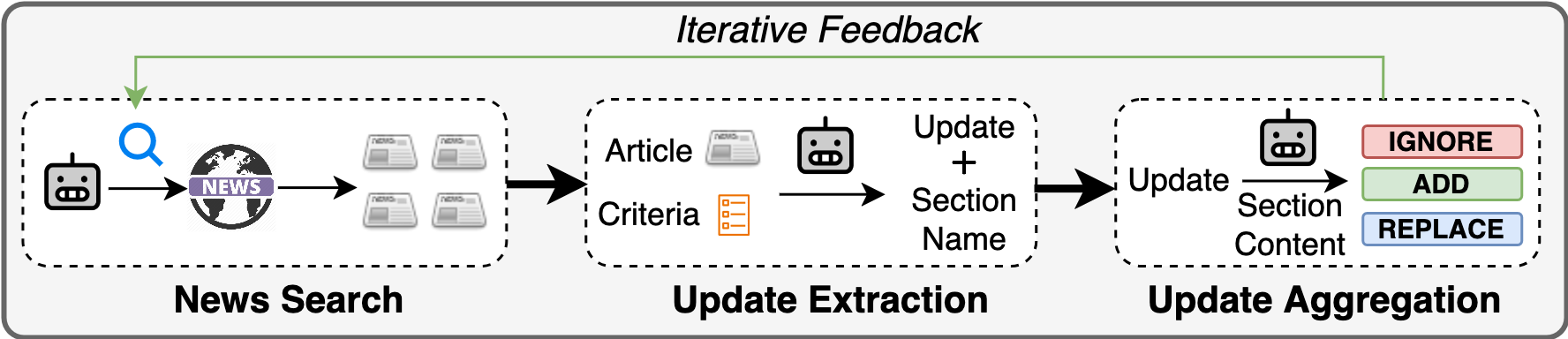}
    \vspace{-1.5em}
    \caption{\name{}'s agentic update aggregation component iteratively performs web searches, identifies potential updates based on section criteria, and aggregates them by deciding whether to ignore, add, or replace existing content, incorporating this feedback to further refine the search process in subsequent steps.}
    \label{fig:update_aggregation}
\end{figure}

The agentic update aggregation process is based on \textit{adaptive information seeking}--rather than using a fixed set of search queries to identify relevant updates, an agent continually assesses which facets of the given article could change, formulates targeted search queries, and dives deeper about new updates as they get identified. Specifically, the aggregation framework, adapted from INFOGENT~\cite{reddy2024infogent}, involves three core components--Navigator, Extractor and Aggregator. At each iteration, the Navigator searches for online sources and identifies a relevant article. The Extractor then extracts the relevant  updates from it along with identifying which section they are relevant to (based on the section criteria from ~\S{\ref{sec:wiki_criteria}}). Finally, the Aggregator leverages the corresponding section content to decide whether the update is worthy being included into the given section, while also accounting for updates aggregated in previous iterations. Figure \ref{fig:update_aggregation} demonstrates this process.

Importantly, the aggregation step also provides \textit{iterative feedback}: if the update extracted from the online source is deemed insignificant or duplicate, the Navigator refines its query to look for updates relating to other aspects of the entity and repeats the process, thereby adaptively closing remaining information gaps. By collapsing redundant suggestions and emphasizing the most salient facts, the Agentic Update Aggregation ensures that downstream fine-grained editing (in \S{\ref{sec:editing}) operates on a concise, coherent set of actionable updates, maximizing the precision of \name{}'s edit recommendations.

\subsection{Fine-Grained Editing}
\label{sec:editing}

Wikipedia, which servers as a continuously evolving knowledge repository, has an extensive history of human editing. These manual edits, which reflect the integration of updated information from external sources, capture human preferences and strategies in updating factual content. Leveraging this resource (details in \S{\ref{sec:human_edits}), we aim to train an editing model capable of integrating new information into Wikipedia articles in a manner that aligns with human editing behaviors. 
This requires preserving factual accuracy, filtering out subjective or irrelevant content while maintaining coherence with the existing content. 

Given historical human edits, we apply filtering (details in Appendix~\S{\ref{sec:editor_filtering}}) to construct our training dataset, with each edit record consisting of three components: (1) the original Wikipedia paragraph, (2) the updated paragraph after editing, and (3) the online source that potentially motivated
the edit. The source content usually includes key factual details that align with the Wikipedia update, along with commentary and subjective elements (details in Appendix~\S{\ref{sec:editor_augmentation}}) acting as noise, simulating real-world reporting. We leverage this data to finetune our editor.  Given an identified update (from~\S{\ref{sec:update_aggregation}}) and the corresponding wiki section paragraph, the editor outputs a fine-grained edit incorporating the update into the section content. 


\section{Data Collection and Evaluation Setup}
\label{sec:data_collection}

Evaluating the edits generated by \name{} poses a significant challenge, as large-scale manual verification is not practical. Consequently, we introduce an automatic evaluation setup utilizing historical human edits as a proxy for ground truth. Our evaluation assesses \name{}'s ability to replicate human-incorporated updates in a historical time period. Specifically, we compare \name{}'s suggested edits, derived from sources published within a defined period ($T$ to $T+\Delta t$) for a Wikipedia article version at time $T$, against actual human edits from the same period. This involves two main steps: (1) extracting historical human edits from Wikipedia articles (\S{\ref{sec:human_edits}}) and (2) mapping these to \name{}'s edits to measure coverage (\S{\ref{sec:automatic_eval}}). Evaluation results are provided in~\S{\ref{sec:end_to_end_results}}.


\subsection{Extracting Human Edits}
\label{sec:human_edits}

Human editing data are obtained by collecting article revision histories within a specific timeframe and identifying modifications between consecutive versions. Edits, extracted by comparing corresponding sections, are categorized as insertions or removals, noting the involved sentences and their paragraphs (details in Appendix~\S{\ref{sec:appendix_human_edits}}).

\paragraph{Identifying Factual Updates:} Many human edits in Wikipedia involve superficial alterations like sentence reordering or formatting changes rather than factual content updates. Thus, filtering steps are applied to retain factual edits involving additions, deletions, or content updates, which are relevant to our evaluation. Subsequently, pinpointing knowledge edits that correspond to information updates involves identifying the addition of new citation URLs, as these typically accompany the integration of new facts by human editors.  These URLs and their source publication dates are collected to assess information recency and edit timeliness via the lag between source publication and human edit timestamp.


\subsection{Automatic Evaluation}
\label{sec:automatic_eval}

\begin{figure}
    \centering
    \includegraphics[width=\linewidth]{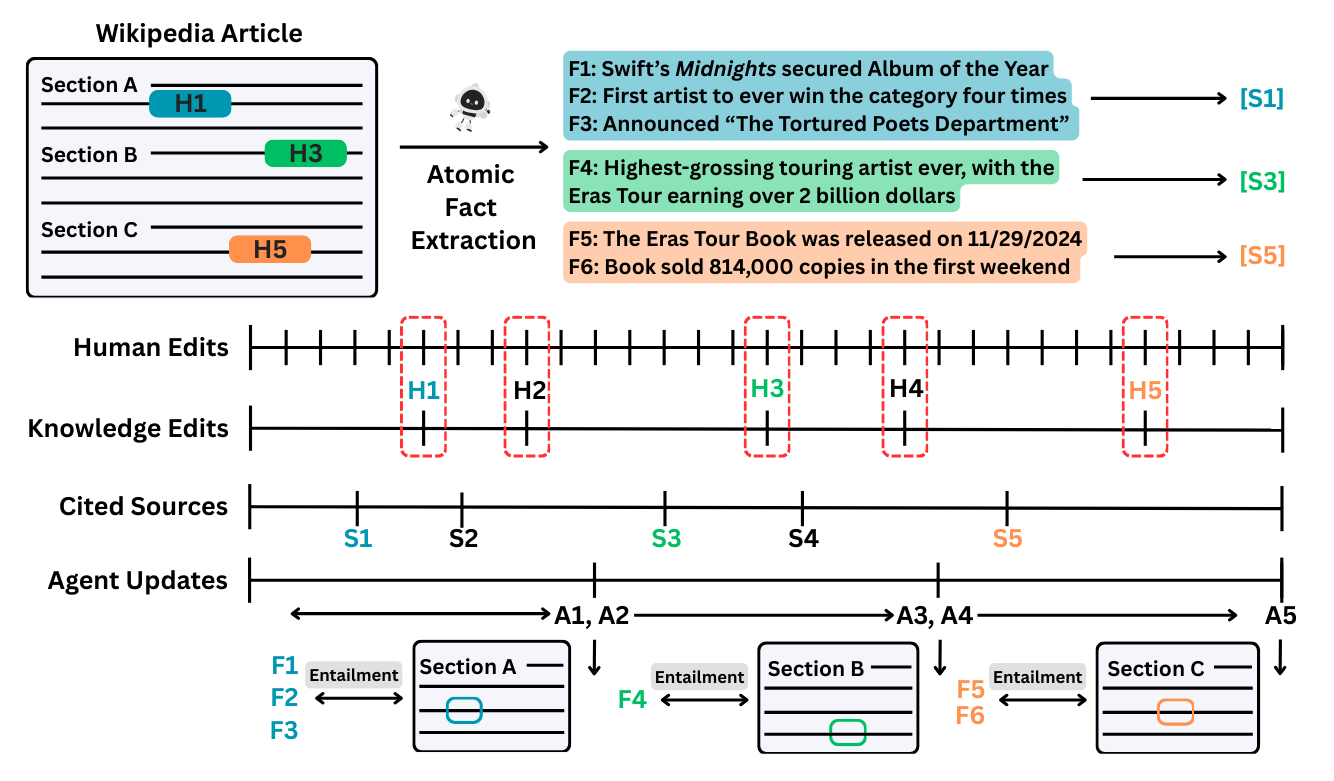}
    \caption{Overview of our automatic evaluation setup comparing agent-generated updates against factual human edits occurring within the same timeframe. By decomposing human edits into atomic facts, we score for coverage by measuring the extent to which agent updates entail these atomic facts.}
    \label{fig:automatic_eval}
\end{figure}

Our automatic evaluation assesses \name{}'s capacity to replicate the factual updates deemed relevant by human editors within a specific interval. Given a Wikipedia article $W$ at time $T$, \name{} proposes edits $E_A = \{e_{a,1}, e_{a,2},...,e_{a,m}\}$ based on sources published within $\Delta t$. Concurrently, actual human edits $E_H = \{e_{h,1}, e_{h,2}, ..., e_{h,n}\}$ applied during $\Delta t$ are considered for comparison.


The primary metric is the coverage of factual human updates by \name{}'s suggestions. Human edits $E_{H}$ are first filtered to a subset $E_{H,\text{factual}}\subseteq E_H$ representing factual updates. The objective is to determine the proportion of edits in $E_{H,\text{factual}}$ semantically matched by an edit in $E_A$.

However, in practice, obtaining the mapping between human and automatic edits is more of a soft matching problem, since a human edit $e_h$ can include multiple pieces of factual information which can be covered by multiple agent edits in $E_A$. Hence, we instead measure human edit coverage based on the presence of atomic facts within them against the agent edits. Each $e_h \in E_{H,\text{factual}}$ is decomposed (via GPT-4o) into atomic facts $F(e_h)=c_1, c_2,....c_k$, where each $c_i$ represents a minimal, verifiable piece of information introduced or modified by $e_h$. The coverage of an atomic fact $c \in F(e_h)$ by $E_A$ is determined using a textual entailment function, $\text{Entail}(c, e_a)\in \{0,1\}$. The coverage status for $c$ is:
\begin{equation*}
    \text{Coverage}(c, E_A) = max_{e_a \in E_A} \text{Entail}(c, e_a)
\end{equation*}

The overall coverage score for a human edit $e_h$ is calculated as the proportion of its constituent $k$ atomic facts that are covered by the agent edits $E_A$:
\begin{equation*}
    \text{Score}(e_h, E_A) = \frac{1}{|F(e_h)|}\sum_{c \in F(e_h)}\text{Coverage}(c, E_A)
\end{equation*}

Further, we define two variants of the coverage metric, \textit{Hard Coverage} and \textit{Soft Coverage}, based on which agent edits are considered for comparison. Let $S(e)$ be the specific section (or subsection/subsubsection) within the Wikipedia article $W$ where an edit $e$ (either human or agent) is applied. \textbf{Hard Coverage} imposes a strict locality constraint, comparing a human edit $e_h$ only against agent edits applied to the \textit{same section}. Specifically, we define the relevant agent edits as $E_{A,S(e_h)} = \{e_a \in E_A | S(e_a) = S(e_h)\}$. 
The overall Hard Coverage, $C_{\text{hard}}$, is defined as: 
\begin{equation*}
    C_{\text{hard}} = \frac{1}{|E_{H,\text{factual}}|}\sum_{e_h \in E_{H,\text{factual}}} \text{Score}(e_h, E_{A,S(e_h)})
\end{equation*}

 On the other hand, \textbf{Soft Coverage} offers a more relaxed evaluation, comparing $e_h$ against all agent edits $E_A$ within $\Delta t$, regardless of edit location:
 \begin{equation*}
    C_{\text{soft}} = \frac{1}{|E_{H,\text{factual}}|}\sum_{e_h \in E_{H,\text{factual}}} \text{Score}(e_h, E_A)
\end{equation*}
 
 This distinction allows us to measure both \name{}'s ability to place information correctly (Hard) and its overall capacity to capture relevant facts (Soft). Figure~\ref{fig:automatic_eval} gives a visual representation of our automatic evaluation setup.



\section{Experiments}
\label{sec:experiments}

We aim to investigate the extent to which \name{} captures updates made by human editors. Our experimental methodology first involves a controlled evaluation of the editing model to measure its ability to incorporate factual updates into the designated section content (\S{\ref{sec:editor_evaluation}}). Subsequently, we assess the end-to-end performance of \name{} in identifying relevant updates and positioning them accurately within the Wikipedia article (\S{\ref{sec:end_to_end}}).

\subsection{Editor Evaluation}
\label{sec:editor_evaluation}
We evaluate the editing model's ability to integrate source information in a manner consistent with human editing behaviors.

\subsubsection{Setup}

\paragraph{Editor Test Data Construction:} For evaluation, we select 600+ entities to create a diverse test set. From each entity, a single edit instance is randomly chosen to avoid any overlap in content. Each data point comprises the original Wikipedia paragraph, edited content and the corresponding online source. Further, every instance is annotated by GPT-4o (prompts in Appendix~\ref{sec:editor_augmentation}) with two key attributes: 
\begin{itemize}[topsep=5pt, partopsep=-5pt, leftmargin=12pt]
    \item \textbf{Key Facts:} Objective facts appearing in both the online source and the final Wikipedia edit but absent in the original paragraph.
    \vspace{-0.5em}
    \item \textbf{Commentary Information:} Subjective or opinionated content from the online source that does not appear in either the original or the revised Wikipedia paragraph, acting as noise.
\end{itemize}

\paragraph{Evaluation Metrics:} Our evaluation is grounded in principles that align with human editing behavior. Specifically, editors tend to make minimal but necessary changes, preserving as much original content as possible while ensuring factual correctness. 
These considerations inform our metrics:
\begin{itemize}[topsep=5pt, partopsep=-5pt, leftmargin=12pt]
    \item \textbf{Token Change:} Number of modified words between the original and updated paragraphs. Lower value indicates a model’s ability to make minimal yet effective edits.
    \vspace{-0.5em}
    \item \textbf{Key Facts Coverage:} Percentage of essential factual information from the online source that is successfully incorporated into the updated paragraph. Higher score reflects proficiency in identifying and integrating crucial updates.
    \vspace{-0.5em}
    \item \textbf{Commentary Information Coverage:} Proportion of commentary content added from the online source. Lower value signifies stronger ability to filter out subjective noise.
\end{itemize}

\begin{table}[t]
    \centering
    \small
    \setlength\tabcolsep{2pt}
    \setlength\extrarowheight{2pt}
    \resizebox{1.0\linewidth}{!}{

    \begin{tabular}{l c @{\hskip 8pt} c @{\hskip 8pt} c}
    
        \toprule
        
        \multirow{2}{*}{\multirow{2}{*}{\textbf{Model}}} & \multirow{2}{*}{\textbf{\makecell{Token\\Change$^\downarrow$}}} &  \multicolumn{2}{c}{\textbf{Information Coverage (\%)}} \\

        \cmidrule(lr){3-4}
        
        ~ & ~ & \textbf{Key Facts$^\uparrow$} & \textbf{Commentary$^\downarrow$} \\
        
        \addlinespace[2pt]
        \midrule
        \addlinespace[2pt]

        \textit{Qwen2.5-7B-Instruct} & 110.1 & \textbf{95.1} & 86.0\\
        \textit{Llama-3.1-8B-Instruct} & 59.1 & 80.0 & 32.5\\
        
        \addlinespace[2pt]
        \midrule
        \addlinespace[2pt]
        
        \textit{GPT-4o} & 73.4 & 91.3 & 53.1\\
        \textit{GPT-4o-mini} & 69.5 & 88.2 & 46.0\\

        \addlinespace[2pt]
        \midrule
        \addlinespace[2pt]
        
        \textit{Qwen2.5-7B-Editor} (ours) & 52.8 & 90.7 & 20.1\\
        \textit{Llama-3.1-8B-Editor} (ours) & \textbf{49.8} & 91.7 & \textbf{18.7}\\

        \addlinespace[2pt]
        \toprule
        \addlinespace[2pt]

        \textbf{Human} (Ideal) & 62.9 & 100.0 & 0.0\\
        
        \bottomrule
    \end{tabular}
    }    
    \vspace{-0.5em}
    \caption{
    Our finetuned editing models outperform their base instruct and GPT counterparts, by using fewer tokens to make edits while retaining key information, and omitting commentary details.
    }
    \label{tab:editing_results}
    \vspace{-1em}
\end{table}

\subsubsection{Results}

Table~\ref{tab:editing_results} presents results from finetuning two state-of-the-art open-source instruction-following models, Qwen2.5-7B-Instruct and Llama-3.1-8B-Instruct. We compare our finetuned models against both the zero-shot baselines as well as closed-source models, specifically GPT-4o and GPT-4o-mini. We highlight key insights as follows:

\paragraph{Raw Qwen models overfit by copying rather than editing effectively:}  
While the raw \texttt{Qwen2.5} model achieves high key information coverage, it does so by excessively copying content, leading to increased inclusion of commentary information. This suggests that a well-calibrated editing strategy is needed, with simply maximizing recall being insufficient as it sacrifices editorial precision.

\paragraph{Our editing models surpass closed-source models, including GPT-4o:}  
Despite having significantly fewer parameters, our models outperform GPT-4o across all key metrics, demonstrating that model scale alone does not guarantee superior editing quality. The zero-shot instruct variants, including GPT-4o, tend to retain excessive commentary, which can introduce bias. Incorporating human editing data via finetuning helps refine the balance between informativeness and neutrality, a key requirement for Wikipedia editing.


\paragraph{Our models have human-level coverage while making fewer token changes:}  The finetuned editor models make minimal yet impactful edits (based on token change), preserving essential information while avoiding unnecessary modifications. Remarkably, our models retain key information at near-human levels while making even fewer modifications. This suggests that, from training on human edit records, the model has learnt an efficient editing strategy. However, opportunities remain for further reducing subjective commentary, to bring it even closer to the ideal standard of neutrality and precision from experienced human editors.

\subsection{End-to-End Evaluation}
\label{sec:end_to_end}

In our end-to-end evaluation, we quantitatively measure how effectively \name{} identifies and incorporates relevant updates from online sources, mirroring the collective factual edits of human editors over a period $\Delta t$. This assesses the system's practical utility in keeping Wikipedia articles up-to-date based on sources published online. 

\subsubsection{Setup}
\label{sec:end_to_end_setup}
\paragraph{Test Set:} The evaluation period spanned from Jan-Dec 2024, with the agent run for individual time periods ($\Delta t$) of two weeks each. We select 45 popular Wikipedia pages, which collectively received over 1400 factual human edits in 2024. To ensure sufficient historical activity for meaningful comparison, only pages with at least 25 human edits within the evaluation period are included.
\paragraph{\name{} Configuration:} Our framework is configured as follows. For Section Ontology Induction (\S{\ref{sec:wiki_criteria}}), which involves understanding page structure and content requirements, GPT-4.1 is utilized as the underlying LLM. The Agentic Update Aggregation step (\S{\ref{sec:update_aggregation}}), responsible for discovering online sources and extracting updates, employs the more efficient GPT-4.1 mini model. The agent's online search capability was powered by the Google Search API, with results restricted to news articles published within $\Delta t$. For each selected Wikipedia page, \name{} identified updates and generated edits using the \texttt{Llama-3.1-8B-Editor}. 
\paragraph{Baselines:} We compare \name{} against two ablations: 1) Utilizing only section names instead of the detailed section-specific criteria derived from \S{\ref{sec:wiki_criteria}}; and 2) Employing a single search query for identifying updates to the article, in contrast to the iterative, multi-query agentic process detailed in \S{\ref{sec:update_aggregation}}. Additionally, we include an oracle baseline, which directly uses URLs cited in human edits as sources. This oracle measures the efficacy of extracting the relevant updates and identifying the correct section to incorporate these updates, assuming perfect source discovery by the agent.


\paragraph{Evaluation Metrics:}Performance is evaluated using the proposed human edit coverage metrics, $C_{\text{hard}}$
  and $C_{\text{soft}}$, as defined in \S{\ref{sec:automatic_eval}}. Section accuracy ($S_{\text{Acc}}$) assesses whether agent edits are made in the same sections as their corresponding human edits.


\begin{table}[t]
    \centering
    \small
    \renewcommand{\arraystretch}{1.35}
    \setlength{\tabcolsep}{4pt}
    \begin{tabular}{l|ccc}
    \toprule
      \textbf{Method}   &  $C_{\text{hard}}$ (\%) & $C_{\text{soft}}$ (\%)& $S_{\text{Acc}}$ (\%) \\
      \midrule
      \name{} & \textbf{15.4} & \textbf{34.4} & \textbf{33.2} \\
        - No Section Criteria  & 15.2 & 33.0 & 28.6 \\
        - No Agentic Search & 9.5 & 21.5 & 19.6 \\
      \demph{Oracle (Human sources)} & \demph{30.6} & \demph{62.2} & \demph{41.4} \\
      \bottomrule
    \end{tabular}
    \vspace{-0.5em}
    \caption{Evaluation of \name{} along with ablations of using section criteria and agentic search. \demph{Oracle} assumes perfect discovery, by directly using source URLs cited in human edits for extracting relevant updates.}
    \label{tab:end_to_end}
\end{table}

\begin{figure}[t]
    \centering
    \includegraphics[width=1.0\linewidth]{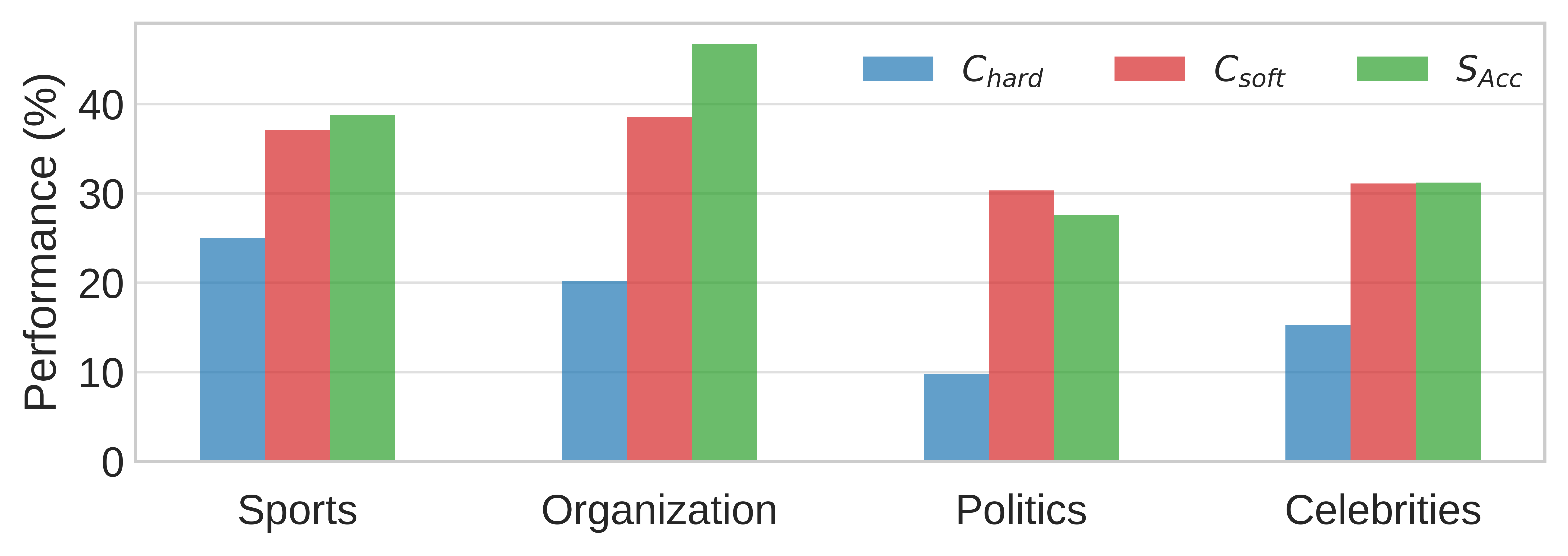}
    \vspace{-1.8em}
    \caption{\name{} performance by page category.}
    \label{fig:category_wise}
    \vspace{-1em}
\end{figure}

\begin{figure*}[t]
    \centering
    \includegraphics[width=1.0\linewidth]{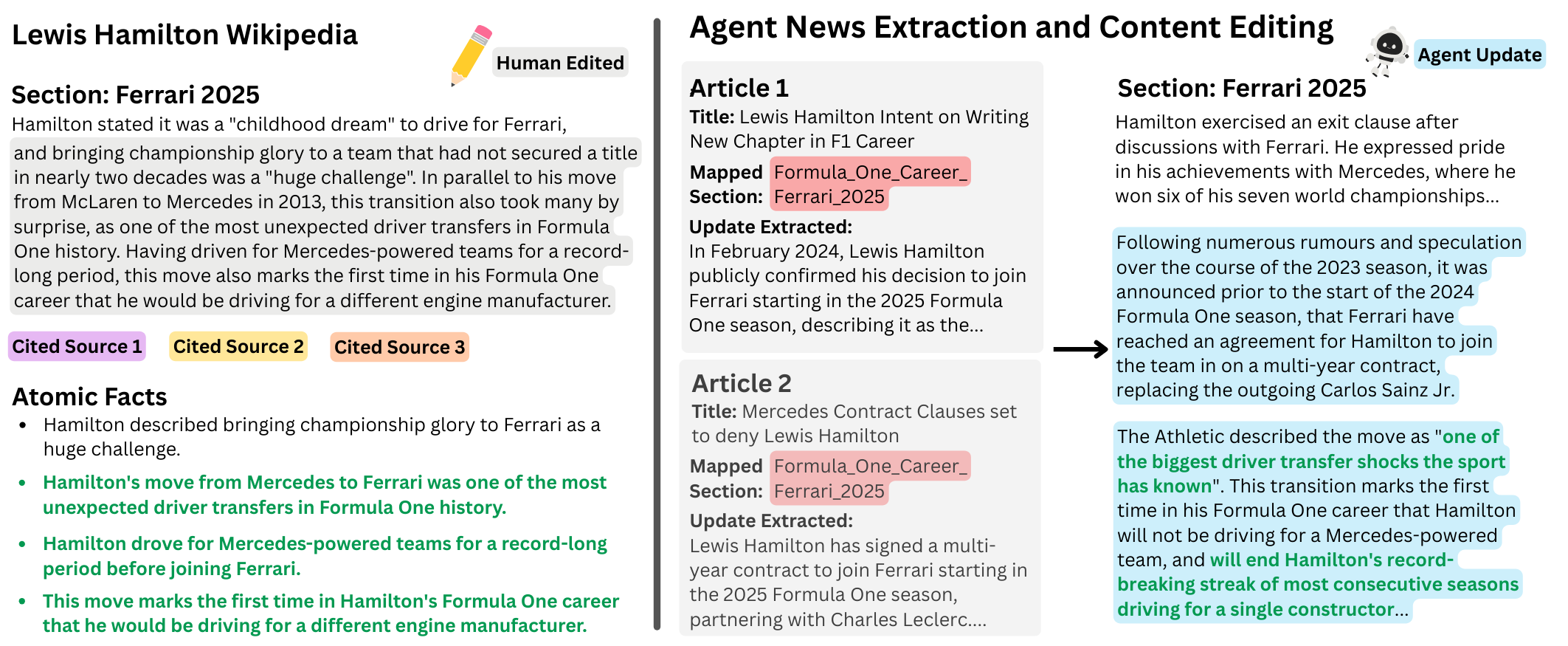}
    \vspace{-1.7em}
    \caption{Qualitative example comparing a human Wikipedia edit (left) with an automated agent update (right) for the `Ferrari 2025' subsection of Lewis Hamilton's page. The agent identified  multiple online sources (center) to generate its update for the correct subsection. The text marked in green points the atomic facts successfully covered from the human edit (bottom left). The agent captured 3 out of 4 atomic facts, resulting in a $C_{\text{hard}}$ score of 0.75.}
    \label{fig:qualitative_example}
    \vspace{-0.75em}
\end{figure*}

\subsubsection{Results}
\label{sec:end_to_end_results}

Table~\ref{tab:end_to_end} presents results, computed as a micro average over the 1400+ factual human edits. \name{} achieves a hard coverage ($C_{\text{hard}}$) of 15.4\% and soft coverage ($C_{\text{soft}}$) of 33.2\%. These results  indicate a substantial capability to automatically incorporate factual information from human edits across diverse pages.
Ablation studies reveal that removing section criteria considerably degrades section accuracy ($S_{\text{Acc}}$), while agentic search is crucial for enhancing coverage. The Oracle, with access to human-cited sources, exhibits markedly higher coverage, yet its $S_{\text{Acc}}$ of 43.7\% underscores the inherent challenge in correctly identifying the target section for updates, even with ground truth sources.


\paragraph{Hard vs. Soft Coverage Insights:} The disparity between $C_{\text{soft}}$ and $C_{\text{hard}}$ (see Appendix Figure~\ref{fig:end_to_end_eval} for page-wise details) is informative. $C_{\text{soft}}$ measures factual content overlap irrespective of placement, whereas $C_{\text{hard}}$ requires the fact to be placed within the same human-edited subsection. $C_{\text{soft}}>C_{\text{hard}}$ suggests \name{} can identify correct factual updates but may struggle in integrating them into the same section as chosen by human editors. This highlights a key challenge in automated Wikipedia editing: determining not only \textit{what} to update but also \textit{where} to integrate it within the existing article. 

\paragraph{Performance by Category:} Performance analysis across four Wikipedia page categories--\textit{sports figures}, \textit{organizations}, \textit{politicians}, and \textit{celebrities}--as shown in Figure~\ref{fig:category_wise}, reveals lower efficacy for politicians and celebrities. We hypothesize that this is due to the high volume of online news coverage for these categories, making it harder to determine which updates are significant and Wikipedia-worthy. Conversely, sports figures and organizations often feature distinct, significant updates, such as sporting events or official press releases.  A qualitative case study (Figure~\ref{fig:qualitative_example}) illustrates a successful update by \name{} on Lewis Hamilton's Wikipedia page, regarding his 2025 Ferrari contract. The system correctly identifies relevant news articles and accurately matches the human edit's subsection. This demonstrates \name{}'s potential for accurate, well-placed edits given clear information and successful section mapping.



\paragraph{Human Evaluation:} To complement automatic evaluations, a human study assessed the acceptability of 100 agent edits. Five experienced Wikipedia editors (each $>$1,000 edits), using the interface in Figure~\ref{fig:eval_ui}, evaluated edits based on common errors and acceptability (accept, accept with revision, reject). Findings indicate 68\% of edits were accepted without needing any changes, 29\% with revisions, and 3\% were rejected. Common issues included stylistic/clarity concerns (17\%), subjective content (6.5\%), and insignificant changes (6.5\%), with most reject decisions corresponding to insignificant changes or irrelevant content (Figure~\ref{fig:eval_results}).

\subsubsection{Discussion}
\label{sec:discussion}
\name{} aims to enhance Wikipedia's timeliness by reducing the latency between information publication and its integration. By automating online update monitoring, \name{} alleviates the burden on human editors, enabling them to focus on verification and quality control. While experiments primarily utilized popular Wikipedia pages to ensure sufficient ground-truth edit data for coverage evaluation, \name{} is expected to be equally applicable, and potentially more impactful, for less popular pages often neglected by human editors. Furthermore, although Wikipedia was chosen for its extensive historical edit data facilitating automatic evaluation, the core agentic aggregation and automatic updating framework of \name{} is generalizable to knowledge bases in other domains.


\section{Conclusion}
This paper introduces \name{}, an agentic framework for autonomously updating Wikipedia articles. Our end-to-end evaluation demonstrates \name{}'s capability to identify relevant updates from online sources and convert them into Wikipedia edit suggestions. While \name{} effectively captures the substance of human edits, evidenced by  $C_{\text{soft}}$, precise alignment with human editorial section placement, reflected in lower $C_{\text{hard}}$,  remains an area for improvement. Future research will target enhancing the agent's section mapping and update integration strategies. We also plan to collaborate with the Wikimedia team to integrate \name{} for the benefit of human editors.


\section*{Limitations}
Wikipedia's reputation for accuracy means any AI-generated content or suggestion must be rigorously reviewed before incorporation. Models that generate or suggest text run the risk of hallucinating--producing plausible-sounding but false statements.
Further, if an AI incorrectly suggests removing sourced content (thinking it's inconsistent or unsupported), it might lead to deletion of valid information. Likewise, automatic text generation could introduce copyright violations if it inadvertently `writes' something too close to a source text, with ongoing discussion about how to attribute AI contributions. Moreover, Wikipedia has strict content policies (neutral point of view, verifiability, no original research) and a specific encyclopedic tone. AI systems can struggle with these nuances. For example, a text generation model might introduce biased language or undue weight without realizing it. Also, there can be resistance or skepticism toward AI suggestions – editors might distrust a black-box recommendation, especially given past issues with bots that made misguided edits. Also, the current version of \name{} mainly edits paragraphs, and cannot update infoboxes or any tables within the articles.
\bibliography{custom}
\clearpage
\appendix

\section{Extracting Human Edits in Wikipedia}
\label{sec:appendix_human_edits}

To obtain authentic human editing data, we extracted edit histories made by human contributors on Wikipedia. For each entity page, we collected all revision versions within a fixed time period and stored them locally. Each revision includes a timestamp, edit details, tags, and other relevant metadata. To identify actual edits, we compared consecutive revisions and extracted the sentences that were modified. 

During data extraction, we observed that some human edits only involve reordering sentences or other superficial changes, without adding, removing, or modifying information. In this study, our goal is to train an agent that can automatically gather and integrate new information into articles. Therefore, our training and testing data must include edits involving actual content changes, rather than simple restructuring. Finally, we filtered out changes involving only punctuation, capitalization, or formatting and focused solely on edits that involved textual or semantic changes.

On Wikipedia, when editors integrate new information, they are typically required to include a source URL as a reference. We use the presence of newly added URLs as an indicator of whether new information has been incorporated. During data extraction, we also collected any new URLs added by editors. Given the URL, we can also obtain the publication date of the referenced source. This date helps us determine the information recency of the edit. The time gap between the source's publication and the human edit timestamp can be used as a measure of the timeliness of human edits. 

In the data collection process, the structure of each collected edit record includes the following information: revision ID, editor, comments, and tags. When extracting text edits, we first segment the raw Wikipedia content by sections. This allows us to obtain the document’s hierarchical structure for each revision, the text content of each section, and any new links added per section. We also detect whether a section contains special elements such as tables, lists, infoboxes, or images.

When comparing revisions, we first check whether the article’s overall hierarchy has changed. If the hierarchy remains the same, we compare the text of each section individually to extract the edits.  However, if the article's hierarchy has changed, we perform a comparison at the full-page level between the two revisions. This section-level edit data is valuable for training the editor.

To simplify the representation of editing behavior, we categorize the editing changes into two types: insertion and removal. This classification effectively captures a human edit as a combination of insertions and deletions, making the data easier to collect and represent.

For each individual human edit, we store the following information: the combination action of insertion and removal actions and the sentences corresponding these actions, and the paragraphs which these sentences belong to, along with the new URLs added as citations.

\section{Editor Training Data Creation}
\label{sec:editor_training_data}
\subsection{Filtering the Edits}
\label{sec:editor_filtering}
To construct a high-quality dataset for training, we first collect Wikipedia edit records corresponding to over 2,000 entities. This dataset includes more than 20,000 human edits along with the original sources that presumably motivated these modifications. To ensure data quality and relevance, we apply a rigorous filtering and refinement process:
\begin{itemize}[topsep=2pt, partopsep=-5pt, leftmargin=12pt]
    \item \textbf{Noise Removal:} We eliminate edits containing unreadable characters, formatting errors, or other forms of noise.
    \item \textbf{Semantic Integrity:} Edits that are excessively long or short, leading to complete rephrasings or drastic changes in paragraph meaning, are discarded.
    \item \textbf{Edit History Simplification:} To avoid complications from excessive back-and-forth changes, we remove instances where the edit history exhibits redundant or conflicting modifications.
    \item \textbf{Section Pruning:} Edits made to non-content sections such as references, external links, or formatting corrections are excluded.
\end{itemize}
Following this rigorous filtering, the dataset is reduced to fewer than 2,000 high-quality edit records suitable for training and testing.

\subsection{Augmenting with Source Content}
\label{sec:editor_augmentation}
While each edit is associated with a citation that may have influenced the modification, many of these sources are either unavailable or contain unreadable content. To address this, we employ GPT-4o to generate plausible source content based on the edits. Specifically, for each edit:
\begin{enumerate}[topsep=2pt, partopsep=-5pt, leftmargin=12pt]
    \item GPT-4o identifies the segment of source content that likely motivated the edit (if available).
    \item We augment this extracted source information into a structured 3-4 sentence paragraph, incorporating:
    \begin{itemize}[topsep=2pt, partopsep=-5pt, leftmargin=12pt]
        \item Key factual details that align with the Wikipedia update.
        \item Commentary and subjective elements acting as noise, simulating real-world reporting.
    \end{itemize}
\end{enumerate}

As a result, each edit record consists of three components: (1) the original Wikipedia paragraph, (2) the updated paragraph after editing, and (3) the augmented source  content that potentially motivated the edit.

\begin{tcolorbox}[
    title=Edit Attributes Annotation Instruction,                 
    colback=blue!5!white,              
    colframe=blue!75!black,            
    fonttitle=\bfseries,               
    sharp corners,                     
    boxrule=1pt,                       
    left=6pt, right=6pt, top=4pt, bottom=4pt, 
    before skip=8pt, after skip=8pt,   
    breakable                          
]
\small
\#\#\# \textbf{Task}\\
You are a helpful assistant to extract the key word or key information from your generated news piece. Please perform the following:\\
1. You should do key word extraction from the news you generated. First, extract key word and phrases that is employed in the modified paragraph given. Try to contain as many key information (date, name, entity, etc.) as possible.\\
2. Next, you should extract those commentary and subjective words and phrases that you added in the news but not employed in the modified paragraph. Try to create a set of words that should not be contained when using the news to update the original paragraph.\\
\\
\#\#\# \textbf{Response Format}\\
\\
\textbf{Original Paragraph}\\
<the original paragraph>\\
\\
\textbf{Modified Paragraph}\\
<the modified paragraph>\\
\\
\textbf{News Piece}\\
<the news piece>\\
\\
Your Response:\\
\\
\textbf{Key Words}\\
<key words and phrases in the news piece that is employed in the modified paragraph>\\
\\
\textbf{Commentary Words}\\
<commentary and subjective words and phrases that you added but should not be contained in the modified paragraph>\\
\\
\#\#\# \textbf{Note}\\
- The key words and phrases extracted should present in both the news piece and the modified paragraph.\\
- The commentary and subjective words and phrases are the ones in news piece but not in the modified paragraph.\\
- All the key words and phrases should be separated by commas.
\end{tcolorbox}

\begin{tcolorbox}[
    title=Editing Instruction,                 
    colback=blue!5!white,              
    colframe=blue!75!black,            
    fonttitle=\bfseries,               
    sharp corners,                     
    boxrule=1pt,                       
    left=6pt, right=6pt, top=4pt, bottom=4pt, 
    before skip=8pt, after skip=8pt,   
    breakable                          
]
\small
You are a helpful assistant to integrate a piece of news information into a Wikipedia article. You should read the original paragraph, find where and how to insert the news information, and return to me a new paragraph with the news information integrated. You should do the following when integrating the news information:\\
1. Only integrate objective news information instead of subjective opinions and commentaries.\\
2. Make less change as possible to the original paragraph.\\
3. Make sure the new paragraph is coherent and grammatically correct.\\
\\
\textbf{Original Paragraph}\\
\{\{Original Content Placeholder\}\}\\
\\
\textbf{News Information}\\
\{\{News Information Placeholder\}\}\\
\\
\textbf{Updated Paragraph}
\end{tcolorbox}

\begin{tcolorbox}[
    title=Evaluation Judgment Instruction,                 
    colback=blue!5!white,              
    colframe=blue!75!black,            
    fonttitle=\bfseries,               
    sharp corners,                     
    boxrule=1pt,                       
    left=6pt, right=6pt, top=4pt, bottom=4pt, 
    before skip=8pt, after skip=8pt,   
    breakable                          
]
\small
\#\#\# \textbf{Task}\\
You are a helpful assistant to judge whether each of the given element is presented in a paragraph. You should judge one by one if it is mentioned in the paragraph and give your reasons. Please follow the instructions below:\\
1. If the exactly same word or phrase appears in the paragraph, then it is considered as mentioned.\\
2. If the word or phrase appears in a different form, such as a synonym or a different tense, but the meaning is the same, then it is also considered as mentioned.\\
3. If the word or phrase does not appear in the paragraph, or the meaning they represent also do not appear, then it is considered as not mentioned.\\
\\
You will be given a list of elements and a paragraph. For each element, you should first give your thought about whether it is mentioned in the paragraph or not based on the standard above. Then you should provide you judgment in "Yes" or "No".\\
\\
\#\#\# \textbf{Response Format}\\
Your Response:\\
- Element: <repeat the first element>\\
- Thought: <your thought>\\
- Judgment: <Yes/No>\\
\\
- Element: <repeat the second element>\\
- Thought: <your thought>\\
- Judgment: <Yes/No>\\
\\
\#\#\# \textbf{Note}\\
- Please make sure your response blocks are in the exact sequence as the elements given. The number of elements given should also match the number of your response blocks.\\
- Please follow the instructions carefully and provide your judgment based on the standard above with thoughtful consideration.
\end{tcolorbox}

\begin{figure*}[t]
    \centering
    \includegraphics[width=1.0\linewidth]{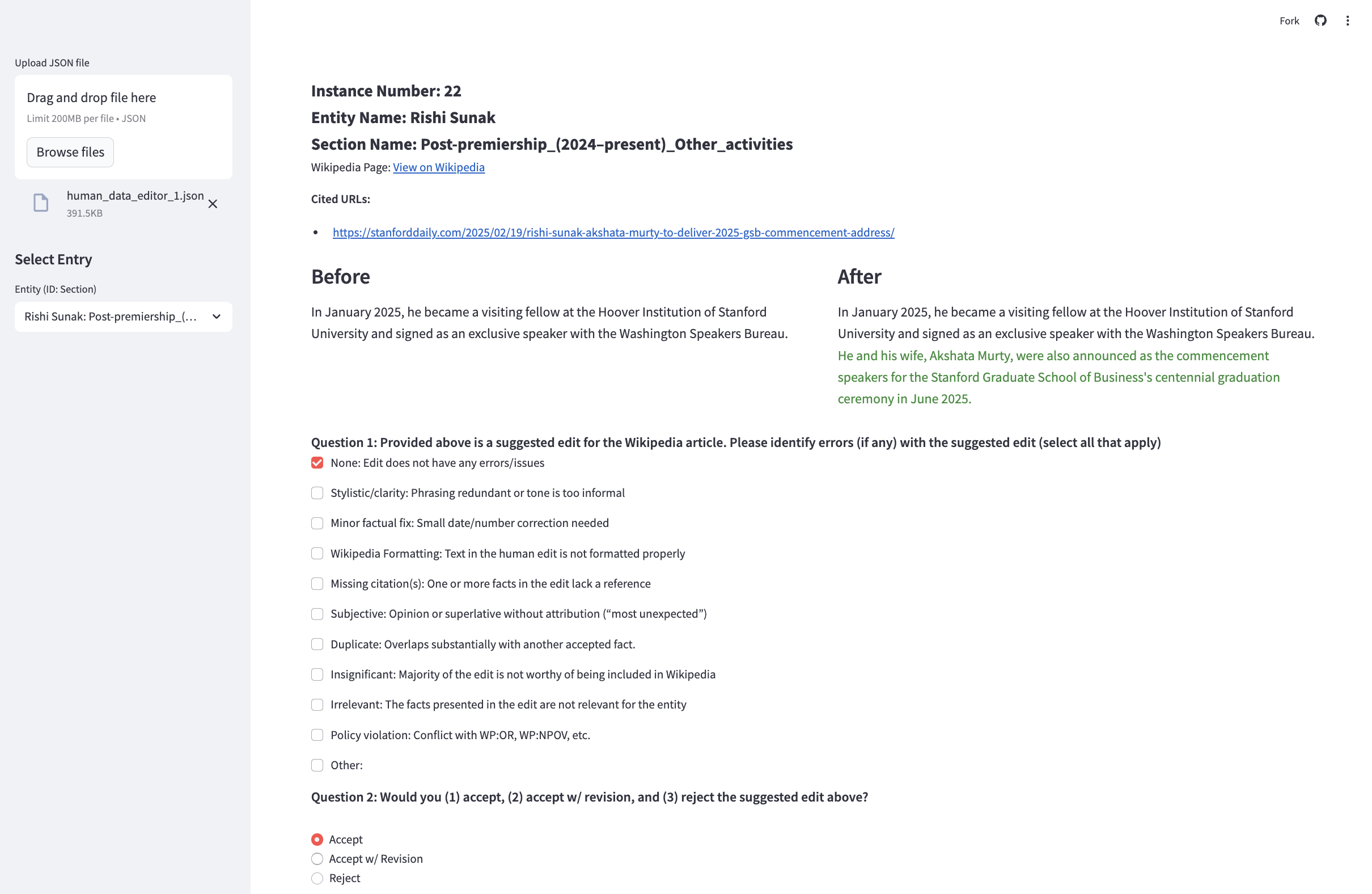}
    \caption{Human annotation task user interface.} 
    \label{fig:eval_ui}
\end{figure*}

\section{Human Evaluation Setup}
In this section, we explain the the setup for the human annotation. 

\subsection{Recruitment}
For our human evaluation, we recruited annotators from local Wikipedia meetups. Wikipedia editors participated in the tasks voluntarily and without compensation. We selected participants who had contributed over 1,000 edits to English Wikipedia and were based in the United States. Our final annotator pool comprised of 5 Wikipedia editors. 

\subsection{Guidelines}
We designed the human annotation user interface to simulate a standard Wikipedia suggestion and acceptance workflow. Therefore, we provided the contextual information that would be normally available which included the (1) the entity name and its corresponding Wikipedia page and (2) the section before and after the suggested edit with the related cited article if necessary

We then asked them to answer two questions surrounding: errors in the suggested edit and acceptance behavior. We designed the first question in collaboration with the Wikipedia editors, collating and de-duplicating a form response. The second question was used to reveal the action in which the Wikipedia editors would take when presented a suggested edit. The annotation user interface can be seen in Figure~\ref{fig:eval_ui}.

\begin{figure*}[t]
    \centering
    \includegraphics[width=1.0\linewidth]{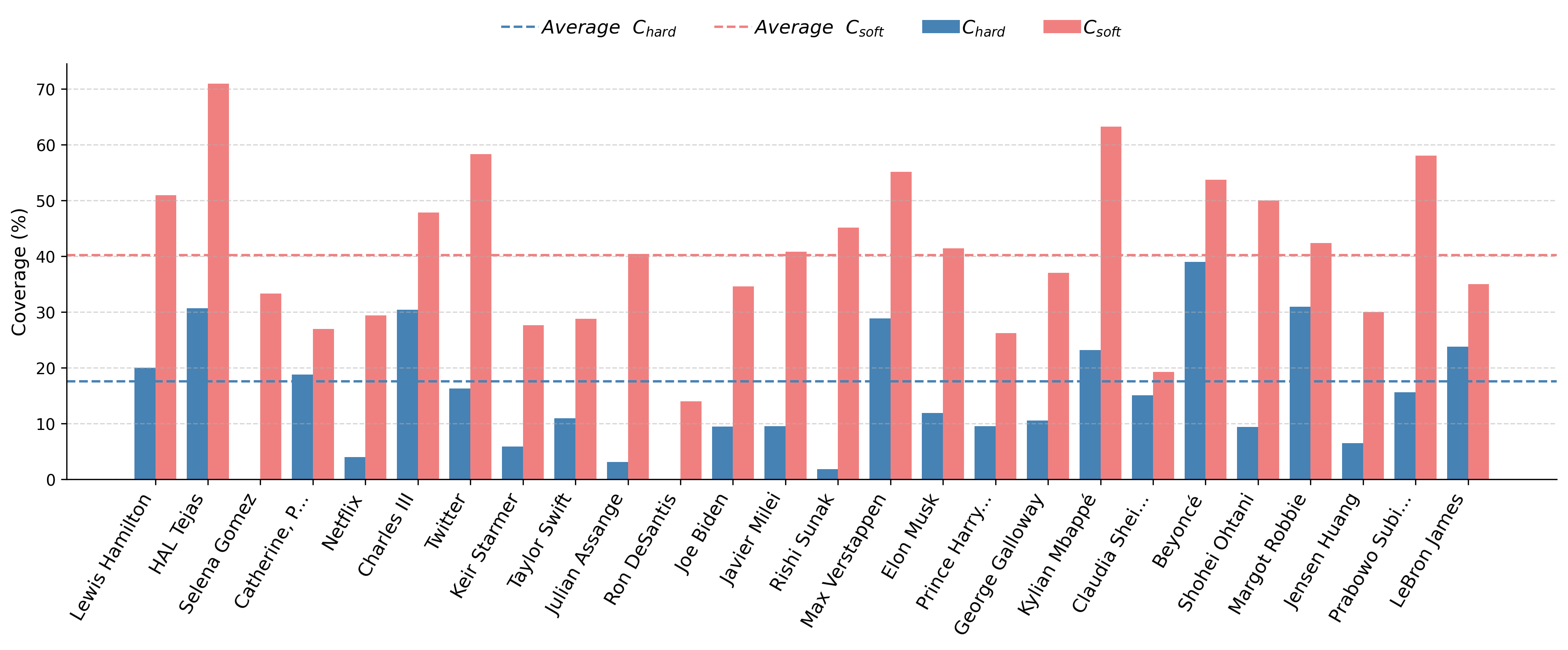}
    \vspace{-2.0em}
    \caption{Human edit coverage scores for \name{} across 20 Wikipedia pages. Both hard and soft coverage metrics (defined in \S{\ref{sec:automatic_eval}}) are shown. Performance varies significantly, with some pages exhibiting low $C_{\text{hard}}$
  (e.g., due to section mismatches) despite factual overlap indicated by higher $C_{\text{soft}}$. Dashed lines represent average scores.}
    \label{fig:end_to_end_eval}
\end{figure*}

\begin{figure*}[!htb]
    \centering
    \includegraphics[width=1.0\linewidth]{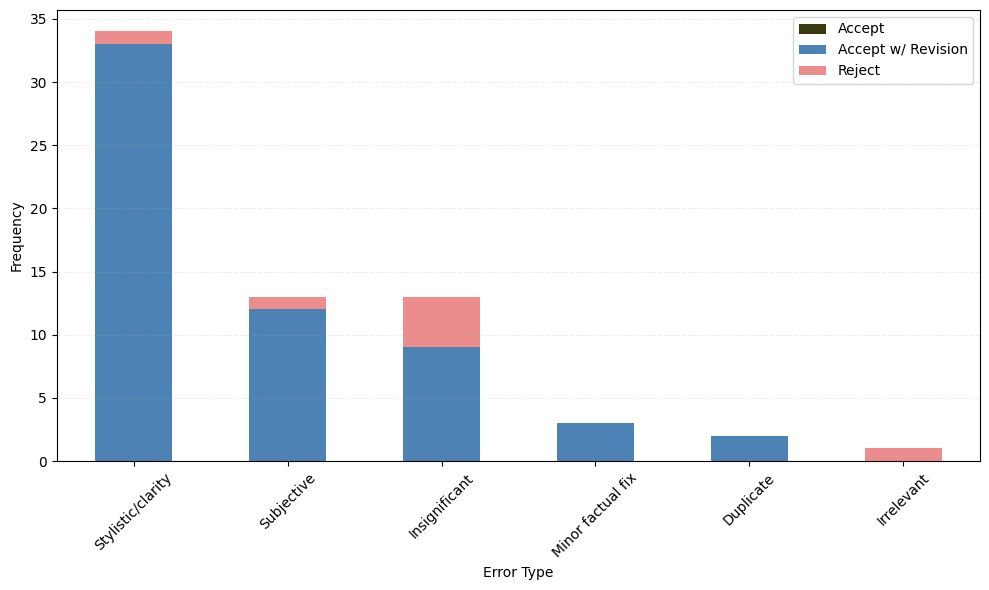}
    \vspace{-2.0em}
    \caption{Results from the human evaluation identifying source errors for suggested edits that would be (1) accepted, (2) accepted with revision, and (3) reject.} 
    \label{fig:eval_results}
\end{figure*}

\end{document}